\documentclass[journal]{IEEEtran}

\usepackage{ifpdf}
\usepackage{cite}
\usepackage[pdftex]{graphicx}
\usepackage[cmex10]{amsmath}
\usepackage{algorithmic}
\usepackage{array}
\usepackage{url}
\usepackage{amssymb}
\usepackage{booktabs}
\usepackage{multirow}
\usepackage[table,xcdraw]{xcolor}
\usepackage[shortlabels]{enumitem}
\usepackage[breaklinks,colorlinks]{hyperref}

\begin{document}

\title{Continual Barlow Twins: continual self-supervised learning for remote sensing semantic segmentation}
%
%
%

\author{Valerio Marsocci, Simone Scardapane
\thanks{Valerio Marsocci is with the Department of Computer, Control and Management Engineering Antonio Ruberti (DIAG), Sapienza University of Rome, 00185, Rome, Italy, {mail: valerio.marsocci@uniroma1.it}}
\thanks{Simone Scardapane is with the Department of Information Engineering, Electronics and Telecommunication (DIET), Sapienza University of Rome, 00184, Rome, Italy, {mail: simone.scardapane@uniroma1.it}}
}

%
%

\markboth{}%
{Marsocci \MakeLowercase{\textit{et al.}}: Continual Barlow Twins}
%



\maketitle

\begin{abstract}
In the field of  Earth Observation (EO), Continual Learning (CL) algorithms have been proposed to deal with large datasets by decomposing them into several subsets and processing them incrementally. The majority of these algorithms assume that data is (a) coming from a single source, and (b) fully labeled. Real-world EO datasets are instead characterized by a large heterogeneity (e.g., coming from aerial, satellite, or drone scenarios), and for the most part they are unlabeled, meaning they can be fully exploited only through the emerging Self-Supervised Learning (SSL) paradigm. For these reasons, in this paper we propose a new algorithm for merging SSL and CL for remote sensing applications, that we call Continual Barlow Twins (CBT). It combines the advantages of one of the simplest self-supervision techniques, i.e., Barlow Twins, with the Elastic Weight Consolidation method to avoid catastrophic forgetting. In addition, for the first time we evaluate SSL methods on a highly heterogeneous EO dataset, showing the effectiveness of these strategies on a novel combination of three almost non-overlapping domains datasets (airborne Potsdam dataset, satellite US3D dataset, and drone UAVid dataset), on a crucial downstream task in EO, i.e., semantic segmentation. Encouraging results show the superiority of SSL in this setting, and the effectiveness of creating an incremental effective pretrained feature extractor, based on ResNet50, without the need of relying on the complete availability of all the data, with a valuable saving of time and resources.
\end{abstract}


%
\IEEEpeerreviewmaketitle

\section{Introduction}
\label{sec:intro}

\begin{figure}
  \includegraphics[width=1\linewidth]{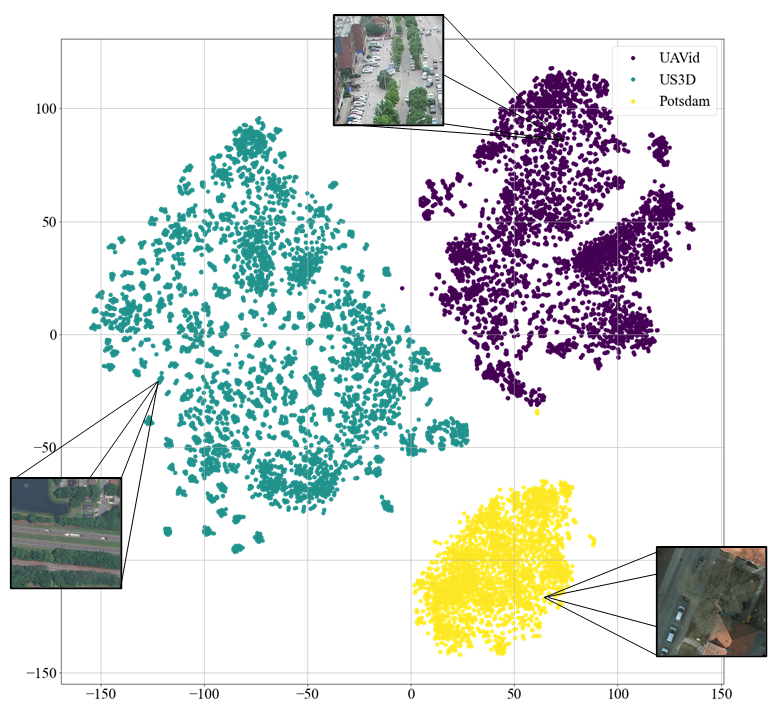}
  \caption{t-Stochastic Neighbor Embedding (t-SNE) visualization results of
the features of three selected RS datasets (see Section \ref{sec:data} for a description of the datasets). It can easily be seen that the images from the different settings can be considered independent but non-identically distributed.}
  \label{fig:tsne}
\end{figure}

\IEEEPARstart{I}{n} recent years, improvements in speed and acquisition technologies have drastically increased the amount of available Earth Observation (EO) images \cite{9705087}. These improvements bring challenging issues to the widespread use of Remote Sensing (RS) classification \cite{9127795} and semantic segmentation techniques, due to (a) the continuous arrival of new data, possibly belonging to partially overlapping domains, (b) the sheer size of the datasets, requiring vast amounts of processing power, and (c) the increasing quantity of data which has not been labeled by a domain expert. The main aim of this paper is to propose an algorithm to deal with these three characteristics simultaneously, that we call Continual Barlow Twins (CBT). This algorithm combines the strengths of two separates lines of research: Continual Learning (CL) for processing a large heterogeneous dataset in an incremental way, satisfying constraints (a) and (b) above, and Self-Supervised Learning (SSL) to deal with the lack of labeling information, satisfying constraint (c) above. In the following, we describe briefly the two issues separately, before introducing our proposed solution.


\subsection*{Problem \#1: EO datasets are heterogeneous}
Consider the situation where we trained a semantic segmentation network on a dataset of Italian satellite images. If we receive a new labeled dataset of similar images from a different nation, ideally, we would like our network to be able to segment equally well images coming from the two countries. However, most of the algorithms developed for classification or segmentation in EO suffer from catastrophic forgetting problems in this context, requiring to discard the acquired knowledge to retrain the model from scratch on the combination of the two datasets \cite{delange2021continual}. In a wide range of EO applications, the strategy of retraining the whole model is computationally expensive and costly \cite{9780164}. Therefore, there is a need to ensure that the newly-developed models have the ability to learn new tasks while retaining satisfying performance on previous ones. The cause of the catastrophic forgetting problem is that different tasks or datasets are independent but not identically distributed in the feature domain, as it is known that the distribution of various RS datasets vary greatly \cite{feng2021continual}, due to different resolutions, acquisitions, textures, and captured scenes. This is even more evident in urban scenes and in datasets made of images acquired from different types of sensors, e.g., drone, airborne, satellite (see Fig. \ref{fig:tsne} for a visualization of this phenomenon). 

In this paper, we leverage a CL algorithm \cite{delange2021continual} to mitigate catastrophic forgetting problem and allow our algorithm to generalize to different feature distributions without the requirement of accessing already seen data.

\subsection*{Problem \#2: EO datasets are largely unlabeled}

Most methods, especially in EO applications, are framed as supervised systems, relying on annotated data. More than in other fields, for drone, aerial and satellite images, it is difficult to rely on a labeled dataset, in light of the high cost and the amount of effort and time that are required, along with domain expertise \cite{9328476}. In computer vision (CV), SSL has been proposed to handle this problem, reducing the amount of annotated data needed \cite{zbontar2021barlow, tian2019contrastive}. The goal of SSL is to learn an effective visual representation of the input using a massive quantity of data provided without any label \cite{chen2020simple}. We can see the task as the need to build a well-structured and relevant set of features, able to represent an image which is useful for several downstream tasks. There is a growing research line demonstrating how SSL techniques increase performance in EO applications \cite{stojnic2018evaluation}, although evaluations have been limited to a single dataset or domain. 

\begin{figure*}
  \centering
  \includegraphics[width=0.9\linewidth]{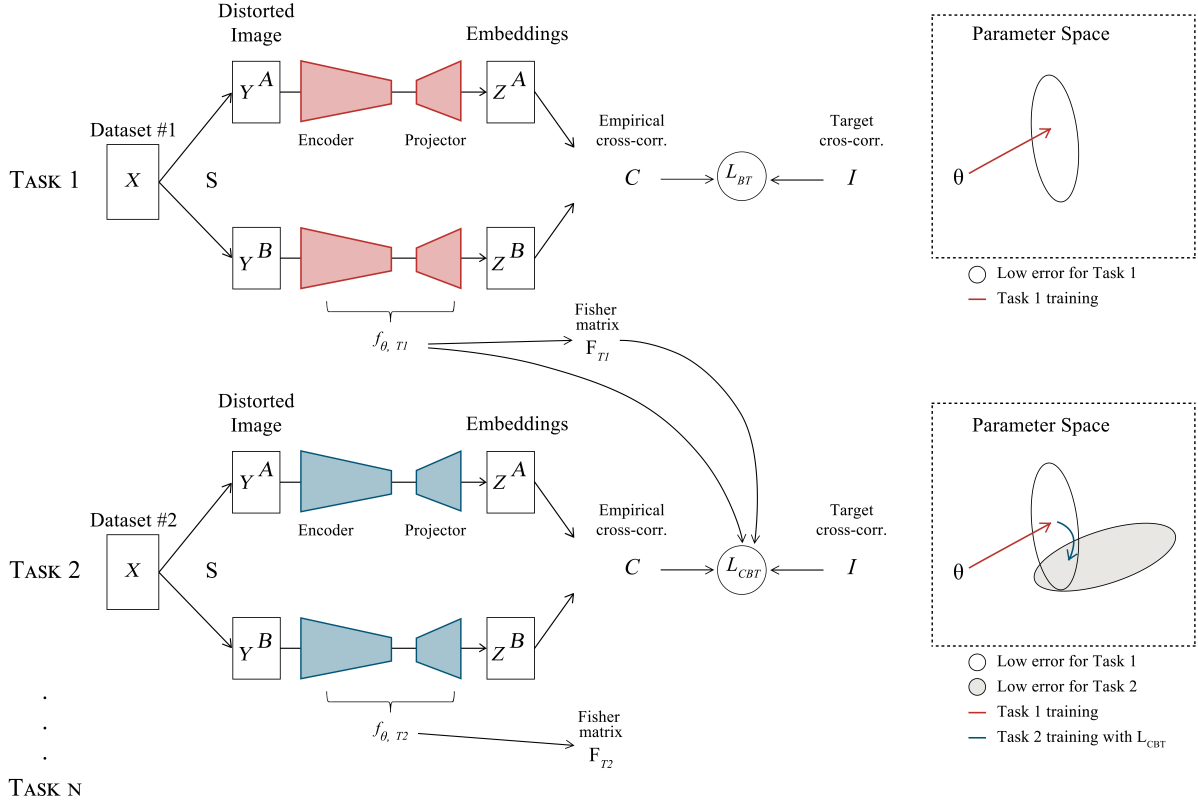}
  \caption{Schematic representation of the Continual Barlow Twins algorithm. When computing $\mathcal{L}_{\mathcal{C} \mathcal{B} \mathcal{T}}$, $C$, and $I$ contribute to the Barlow Twins loss term, $f_{\theta, T1}$, $F$, and $f_{\theta, T2}$ to the $EWC$ regularization term.}
  \label{fig:cbt}
\end{figure*}

\subsection*{Contributions of the paper}

In this paper, we propose and experimentally evaluate a novel algorithm that is able to train a deep network for RS by exploiting vast amounts of heterogeneous, unlabeled, continually-arriving data. Specifically, we show that it is possible to exploit the potential of SSL \textit{incrementally}, to obtain an efficient and effective pretrained model trained in several successive steps, without the need to re-train it from scratch every time new data is added \cite{fini2021self}. The proposed \textit{Continual Barlow Twins} (CBT) algorithm trains a feature extractor (ResNet50) with Barlow Twins (BT) \cite{zbontar2021barlow}, whose loss is integrated with a regularization term, borrowed from Elastic Weight Consolidation (EWC) \cite{kirkpatrick2017overcoming}, to avoid catastrophic forgetting. With the obtained feature extractor, we train a UNet++ \cite{zhou2018unet++} to perform semantic segmentation. When acquiring new RS data, CBT can be trained quickly, as it will be necessary to update it on the new data only, discarding all old data. Our method also provides computational efficiency, potentially allowing small realities to train a large model on huge amounts of data that would be unfeasible otherwise. 

Since the generalisation capabilities and benefits of SSL on RS data from non-overlapping domains (as shown in Fig. \ref{fig:tsne}) is still unexplored, we also propose a new benchmark by combining three datasets with images captured with different sensors (drone, airborne and satellite data), with different resolutions and acquired under different conditions, representing different objects and scenes. We show that SSL targeted to RS images can outperform standard pre-trained strategies (e.g., ImageNet), and we expect this to become a useful benchmark scenario for further research in SSL and CL in remote sensing. We also show that the proposed CBT algorithm offers significantly more versatility and less computing time compared to a standard approach.

\section{Related Works}
\label{sec:relworks}

In this section, we briefly review the relevant literature on SSL (Section \ref{sec:ssl}) and CL (Section \ref{sec:cl}) in EO. 
We note that in CV, the combination of these two technologies is starting to be explored, and some works have already started to demonstrate how SSL methods are well prepared, after small modifications, to learn incrementally \cite{fini2021self, caccia2021special}. On the other hand, in EO, this combination has not yet been explored, to our knowledge, apart from a few embryonic contributions combining weakly-supervision with CL \cite{lenczner2022weakly}.

\subsection{Self-supervised Learning in EO}
\label{sec:ssl}

In \cite{stojnic2021self}, the authors train CMC \cite{tian2019contrastive} on three large datasets both with RGB and multispectral bands. Then, they evaluate the effectiveness of the learned features on four datasets to solve downstream tasks of both single-label and multi-label classification.
The same authors, in \cite{stojnic2018evaluation}, applies a split-brain autoencoder on aerial images.
In \cite{marsocci2021mare}, the authors perform a semantic segmentation downstream task on the Vaihingen dataset \cite{rottensteiner2012isprs} to learn the features of the encoder of the net that solves the segmentation.
A different approach is proposed in \cite{ayush2021geography}, where the authors adopt as contrastive strategy the use of RS images of the same areas in different time frames, introducing a loss term based on the geolocation of the tiles. Other contrastive strategies are proposed by \cite{tao2020remote} and \cite{kang2020deep}. \cite{vincenzi2021color} learns visual representations inferring information on the visible spectrum from the other bands on BigEarthNet \cite{sumbul2019bigearthnet}. 
In \cite{dong2020self}, the authors propose a net that, imitating the discriminator of a generative adversarial network (GAN), identifies patches taken from two temporal images. A similar approach, with multi-view images, is proposed in \cite{chen2021self}. In \cite{9252123}, the authors show the effectiveness of a SSL pretraining for time-series classification. Finally, \cite{9964200} uses SSL strategy for transfer learning super resolution. purposes.

\subsection{Continual Learning in EO}
\label{sec:cl}


In \cite{ammour2021continual}, the authors proposed a two-block network for RS land cover classification tasks, where one module minimizes the error among classes during the new task training, and another module learns how to effectively distinguish among tasks, based on representing past data stored in a linear memory. Similarly, \cite{tasar2019incremental} proposes a framework based on two purposes: adapting and remembering. Concerning the former, the authors save a copy of the trained-on-the-previous task net, to store the info of the already seen classes. The latter stores some old data, which feed the net during the sequential training steps. Shaped for semantic segmentation, \cite{feng2021continual} propose two regularization components: representation consistency structure loss and pixel affinity structure loss. The first retains the information in the isolated pixels. The second saves the high-frequency information, shared throughout the tasks.
%


\section{Methodology}
\label{sec:method}

\subsection{Overview of the components}

We consider a RS scenario where (a) data is coming incrementally from multiple domains (e.g., drone, airborne and/or satellite images); (b) we cannot re-train from scratch the model when new data is received; (c) the majority of the data is unlabeled. We refer to each domain (or subset of the dataset) as a \textit{task}, in accordance with the CL literature. To achieve our compound objective, the intuition is to embed a CL strategy in a self-supervised framework, by combining two algorithms that are considered state-of-the-art in their respective fields: (i) Barlow Twins (BT) \cite{zbontar2021barlow}, which trains a network based on measuring the cross-correlation matrix between the outputs of two identical networks fed with distorted versions of a sample, and making it as close to the identity matrix as possible; (ii) Elastic Weight Consolidation (EWC) \cite{kirkpatrick2017overcoming} consisting of constraining important weights of the network to stay close to the value obtained in previous tasks. In the next section, we highlight in detail how Continual Barlow Twins (CBT) works. A schematic overview of the method is provided in Fig. \ref{fig:cbt}. 

\subsection{Continual Barlow Twins}

Consider for now a single task, and denote by $X$ a batch of unlabeled images. Our main training step, taken from BT, produces two disturbed views of $X$, $Y_A$ and $Y_B$, based on a set of data augmentations strategies $S$ (e.g., random rotations and scalings). In this paper we consider standard sets of data augmentations (see Section \ref{sec:exp}), although augmentations specific to RS could also be considered. The two views are fed to a convolutional neural network with weights $\theta$, that produces, respectively, two embeddings $Z_A$ and $Z_B$ (assumed to be mean-centered along the batch dimension).
To learn effective representations of the input images in a self-supervised fashion we leverage the BT loss, which is composed of two terms called \textit{invariance} and \textit{redundancy reduction} terms:

\begin{equation}
\label{eq:bt}
\mathcal{L}_{\mathcal{B} \mathcal{T}}(X) = \underbrace{\sum_{i}\left(1-\mathcal{C}_{i i}\right)^{2}}_{\text{invariance term }}+\mu \underbrace{\sum_{i} \sum_{j \neq i} \mathcal{C}_{i j}^{2}}_{\text{redundancy reduction term }}
\end{equation}
where $\mu$ is a positive constant balancing the invariant and the redundancy reduction terms of the loss, and $\mathcal{C}$ is the cross-correlation matrix, with values comprised between -1 (i.e. total anti-correlation) and 1 (i.e. total correlation), computed between the outputs of the two identical networks along the batch dimension. Practically, the first term of the loss has the goal to make the diagonal elements of $\mathcal{C}$ equal to 1. In this way, the embeddings become invariant to the applied augmentations. On the other hand, the second term of the loss has the aim to bring to 0 the off-diagonal elements of $\mathcal{C}$. This ensures that the various components of the embeddings are decorrelated with each other, making the information non-redundant, enhancing the representations of the images.

Suppose now that the network has been trained on images coming from a task $T_1$ using the loss \eqref{eq:bt} (e.g., drone images), and we receive a new dataset of images coming from a second task $T_2$ (e.g., satellite images). We denote the weights obtained at the end of the first training as $\theta^{T_1}$, and the data of the two tasks respectevely as $D_{T_1}$ and $D_{T_2}$. To retain old knowledge from $T_1$ and avoid catastrophic forgetting, we complement the BT loss \eqref{eq:bt} with a EWC regularization term \cite{kirkpatrick2017overcoming} which forces the weights to stay close to $\theta^{T_1}$ depending on their importance, given by the diagonal of the Fisher information matrix $F$, which is a positive semidefinite matrix corresponding to the second derivative of the loss near the minimum. In our scenario, the loss cannot be decomposed for each individual data point, as it depends on the cross-correlations between data in a mini-batch and its corresponding augmentations. To this end, denoting by $B_{T_1}$ the number of mini-batches $X$ that can be extracted from $D_{T_1}$, we approximate the $i$-th element of the diagonal Fisher information matrix as:
\begin{equation}
    F_i = \frac{1}{B_{T_1}} \sum_{X \in D_{T_1}} \left[ \frac{\partial \mathcal{L}_{\mathcal{B} \mathcal{T}}(X)}{\partial \theta_i^{T_1}} \right]^2
\end{equation}
where $\mathcal{L}_{\mathcal{B} \mathcal{T}}(X)$ denotes the BT loss computed on mini-batch $X$, as in \eqref{eq:bt}. Intuitively, each weight of the network is given an importance that depends on the square of the corresponding loss gradient. Given this approximation, the new loss for a batch of images $X$ taken from the second task is given by:

\begin{equation}
\label{eq:ewc}
\mathcal{L}(X) = \mathcal{L}_{\mathcal{B}\mathcal{T}}(X)+\sum_{i} \frac{\lambda}{2} F_{i}\left(\theta_{i}-\theta^{T_1}_i\right)^{2}
\end{equation}
where $\mathcal{L}_{\mathcal{B}\mathcal{T}}(X)$ is the BT loss \eqref{eq:bt} computed on the data from task $T2$, and $\lambda$ weights the constraint on the previous task. If moving to a third task, we repeat the computation of the Fisher information matrix at the end of training for the second task and replace it. The CBT approach is summarized in Figure \ref{fig:cbt}, and the associated code is available online.\footnote{\url{https://github.com/VMarsocci/CBT}} After the self-supervised pre-training, the network can be exploited for any downstream task of interest in EO. In particular, we explore in Section \ref{sec:exp} a fine-tuning for a semantic segmentation task.

\section{Datasets}
\label{sec:data}

To perform the experiments we build a novel dataset which is a combination of three previously introduced datasets. Each contains images from a different source: airborne, satellite, and drone. As previously stated, the construction of a novel mixed dataset is crucial since the data is vastly heterogeneous, presenting almost non-overlapping domains, as shown in Fig. \ref{fig:tsne}. In fact, the choice was dictated by the desire to demonstrate the effectiveness of the SSL on a challenging task (that is semantic segmentation), extending its validity even in the case of highly variable data, while most previous works focused on a single domain (see Section \ref{sec:relworks}). We briefly summarize next each dataset. Salient information are summed up in Table \ref{tab:datasets}. 


\begin{table}
  \centering
  \caption{Summary of the datasets used for the experiments.}
  \begin{tabular}{ccc}
    \toprule
    Dataset & Type of images & Number of images\\
    \midrule
    Potsdam \cite{rottensteiner2012isprs} & Aerial & $\sim$5000\\
    UAVid \cite{lyu2020uavid} & UAV & $\sim$7500\\
    US3D \cite{bosch2019semantic} & Satellite & $\sim$11000\\
    \bottomrule
  \end{tabular}
  \label{tab:datasets}
\end{table}

\subsubsection{Potsdam}
\label{sec:dpot}
The ISPRS Potsdam dataset \cite{rottensteiner2012isprs} consists of 38 high-resolution aerial true orthophoto (TOP), with four available bands (Near-Infrared, Red, Green and Blue). Each image is $6000\times6000$ pixels, with a Ground Sample Distance (GSD) of 5 cm, ending up in covering 3.42 $km^2$. For our experiments, we took in consideration only the 38 RGB TOPs. These are annotated with pixel-level labels of six classes: background, impervious surfaces, cars, buildings, low vegetation, trees. We used the eroded mask, and we selected 24 images for training, 13 for testing and 1 for validation, without considering the background class, similarly to \cite{li2021multiattention}. We cropped each image in $512\times512$ non-overlapping patches, ending up in 2640 images for training, 120 for validation and 1680 for test. 

\subsubsection{UAVid}
\label{sec:duav}
Unmanned Aerial Vehicle semantic segmentation dataset (UAVid) \cite{lyu2020uavid} consists in 42 video sequences, captured with 4K high-resolution by an oblique point of view. UAVid is a challenging dataset due to the very high resolution of images, large-scale variation, and complexities in the scenes. The authors extracted ten labelled images per each sequence, ending up in 420 images with $3840\times2160$ pixels. The annotated classes are 8: building, road, static car, tree, low vegetation, human, moving car, background clutter.
The images are already divided into train, validation and test, by the authors, however the test segmentation maps have not yet been released. For this reason, we used a part (80\%) of the validation set as test set in our experiments. Moreover, we cropped the images in $512\times512$ non-overlapping patches, ending up in $\sim7500$ images.

\subsubsection{US3D}
\label{sec:dus}
The US3D dataset \cite{bosch2019semantic} includes approximately 100 $km^2$ coverage for the United States cities of Jacksonville, Florida and Omaha, Nebraska. Sources include incidental satellite images, airborne LiDAR, and feature annotations derived from LiDAR. The dataset is composed of 2783 images, $1024\times1024$, obtained from the WorldView-3 satellite: they are non-orthorectified and multi-view. The images have 8 bands, six of which are part of the visible spectrum, and two of the near infrared. Semantic labels for the US3D dataset, derived automatically from Homeland Security Infrastructure Program (HSIP), are five: ground, trees, water, building and clutter. For our experiments we considered only RGB bands and all the classes. Also for this dataset, we cropped the images in $512\times512$ non-overlapping patches, ending up in more than 11000 images, randomly divided in train ($\sim70\%$), validation ($\sim10\%$) and test ($\sim20\%$).

\section{Experimental Setup}
\label{sec:exp}

For the training phase, a single Tesla V100-SXM2 32 GB GPU has been used.
For the semantic segmentation task, we use UNet++ \cite{zhou2018unet++}, with the Squeeze and Excitation strategy \cite{roy2018recalibrating} and the softmax function as activation on the last layer. For the experiments on all the three datasets, we fix the batch size to 8, the number of epochs to 200 and the learning rate to 0.0001. Moreover, we used Adam as optimizer, Jaccard loss as the cost function and the following set of augmentations: random horizontal flip, random geometric transformation (i.e. shifting, scaling, rotating), random gaussian noise, random radiometric transformation (i.e. brightness, contrast, saturation variations). The mean intersection over union (mIoU), and F1-score (F1) are the selected evaluation metrics. 
Under these conditions, we test different pre-training strategies for UNet++. As baselines, we use ResNet50 pretrained on ImageNet in a supervised manner, and ResNet50 pretrained on ImageNet data with BT\footnote{Downloaded at \url{https://github.com/facebookresearch/barlowtwins}}, considered the fact that there are not other proposed algorithm for these kind of task. For the proposed CBT approach, we pretrain ResNet50 incrementally on the three datasets in this order: US3D, UAVid, Potsdam (with $\lambda=10e-2)$. Then, we left the rest of parameters as in \cite{zbontar2021barlow}. As additional baseline (i.e. upperbound baseline), we consider ResNet50 pretrained on the three datasets with standard BT. We used the same set of parameters provided in \cite{zbontar2021barlow}. Finally, to assess the catastrophic forgetting we trained another encoder trained consequentially on the three datasets (US3D, UAVid, Potsdam) with a vanilla BT, without CL constraints. With these encoders, we trained the semantic segmentation models in a supervised way, with different percentages (10\%, 50\%, 100\%) of labeled data for the three datasets. Particularly, we run all the experiments three times, reporting the mean and the standard deviation of the resulting metrics.
The results are shown and commented in the next Section (Sec. \ref{sec:expres}). 

\section{Experimental Results}
\label{sec:expres}

\begin{figure}
  \includegraphics[width=1\linewidth]{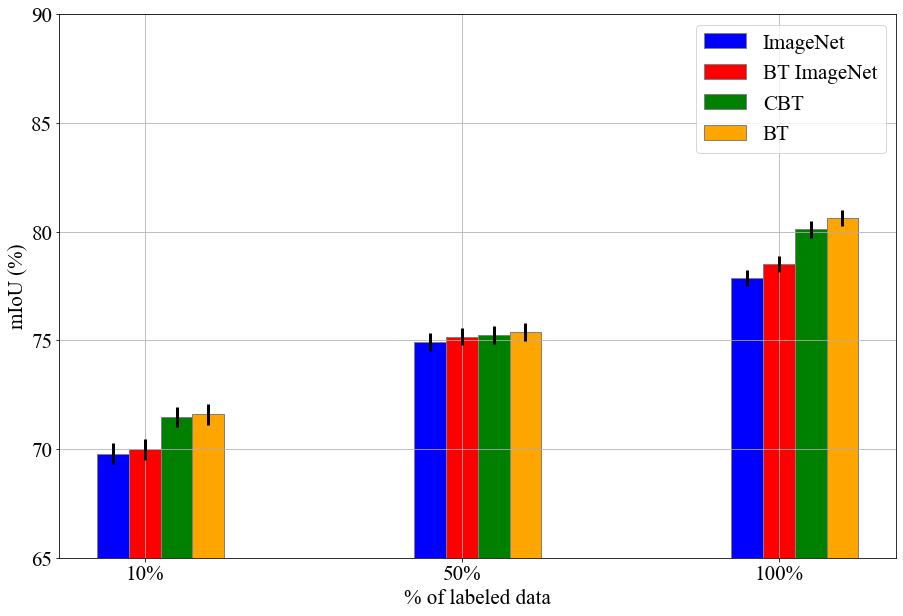}
  \includegraphics[width=1\linewidth]{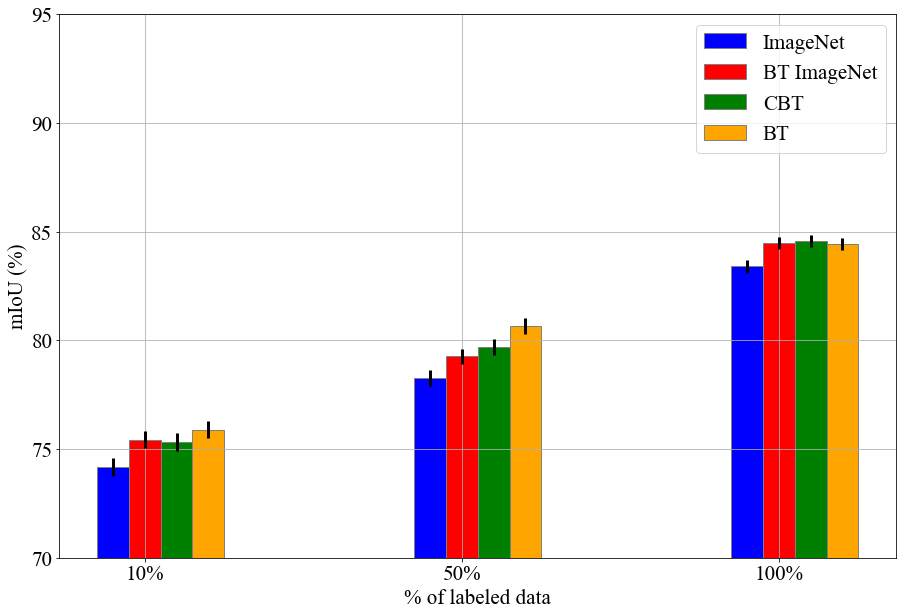}
  \includegraphics[width=1\linewidth]{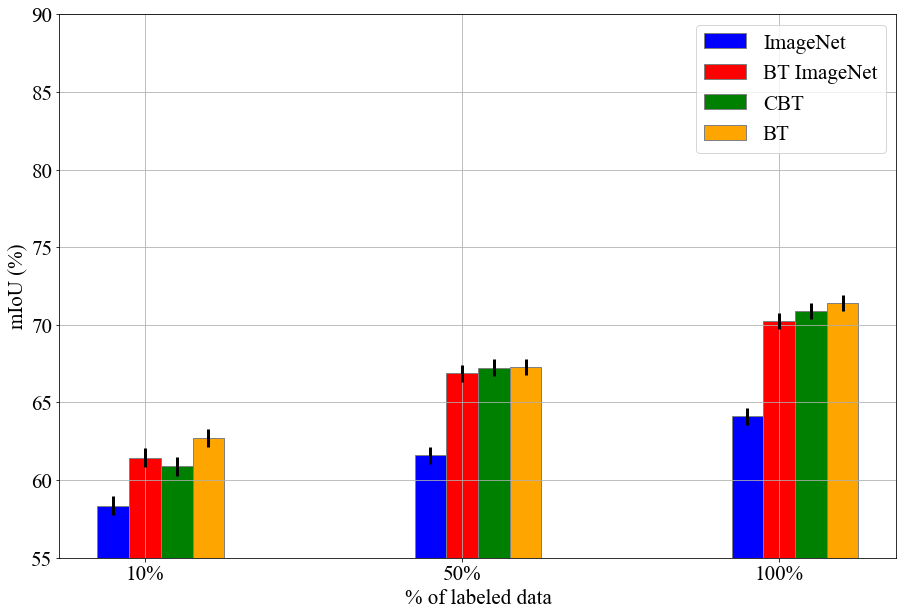}
  \caption{mIoU metrics on experiments with an increasing amount of labeled data of respectevely a) UAVid, b) US3D and c) Potsdam.}
  \label{fig:miou}
\end{figure}

\begin{figure}
  \includegraphics[width=0.85\linewidth]{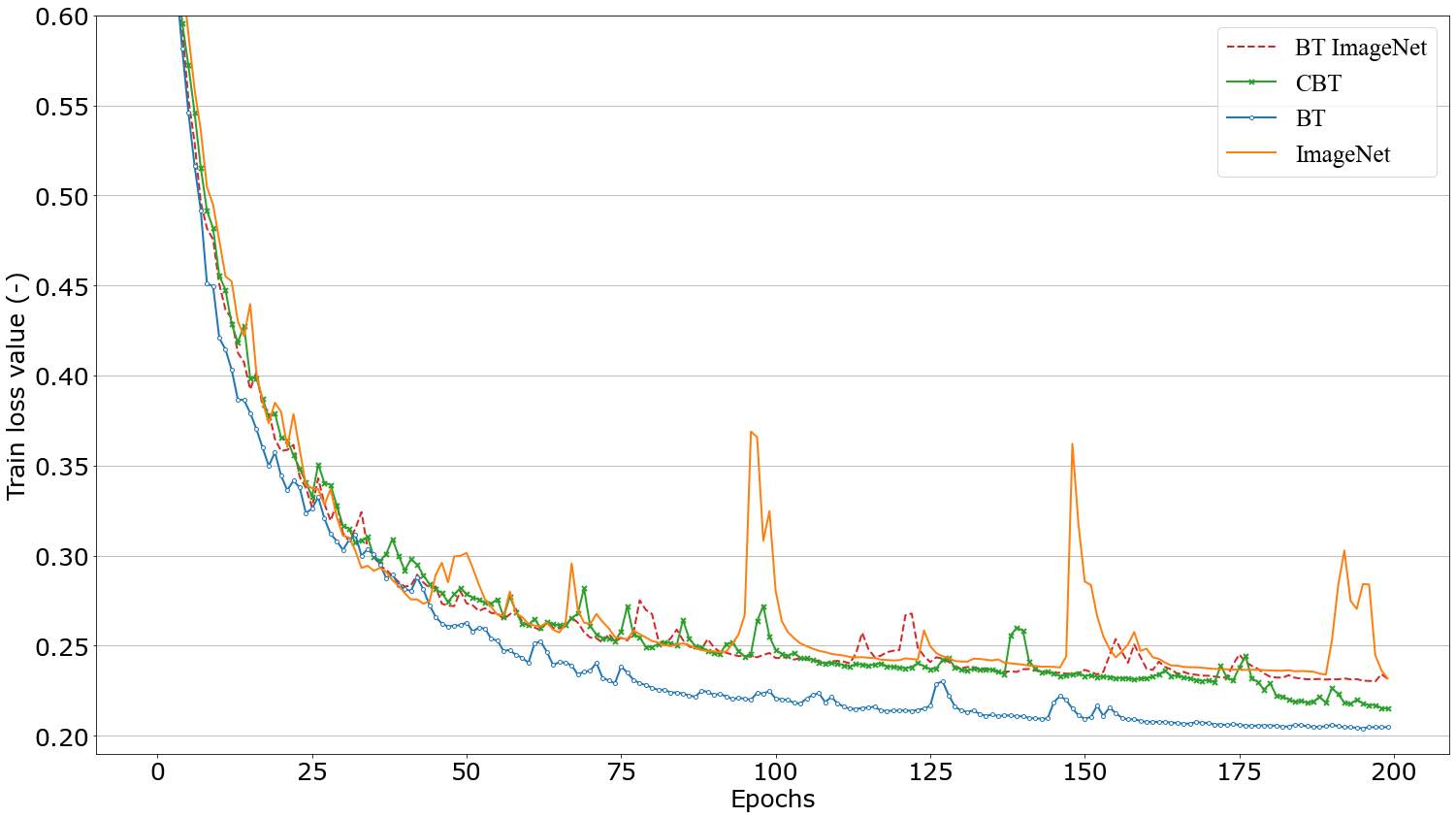}
  \includegraphics[width=0.85\linewidth]{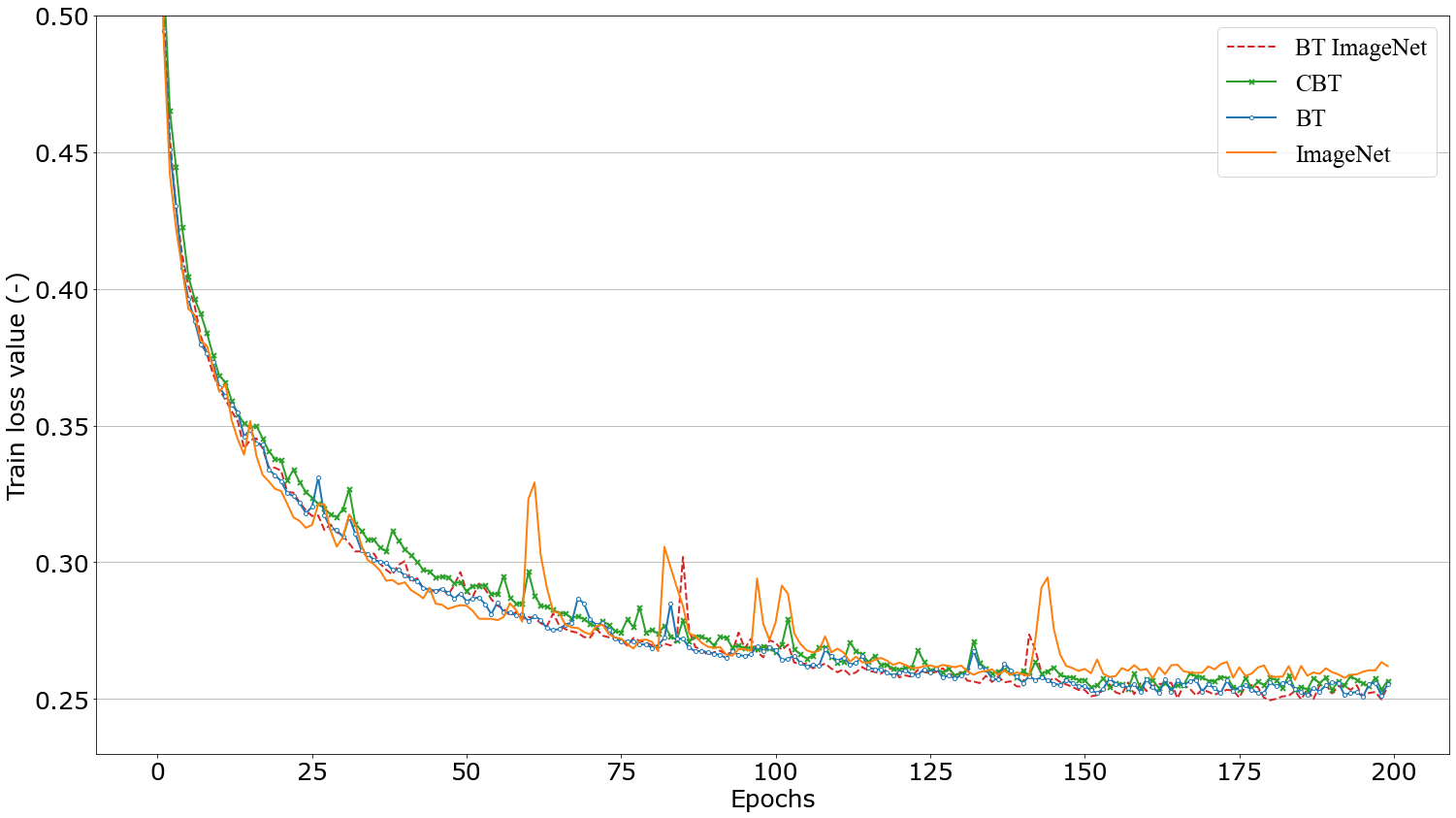}
  \includegraphics[width=0.85\linewidth]{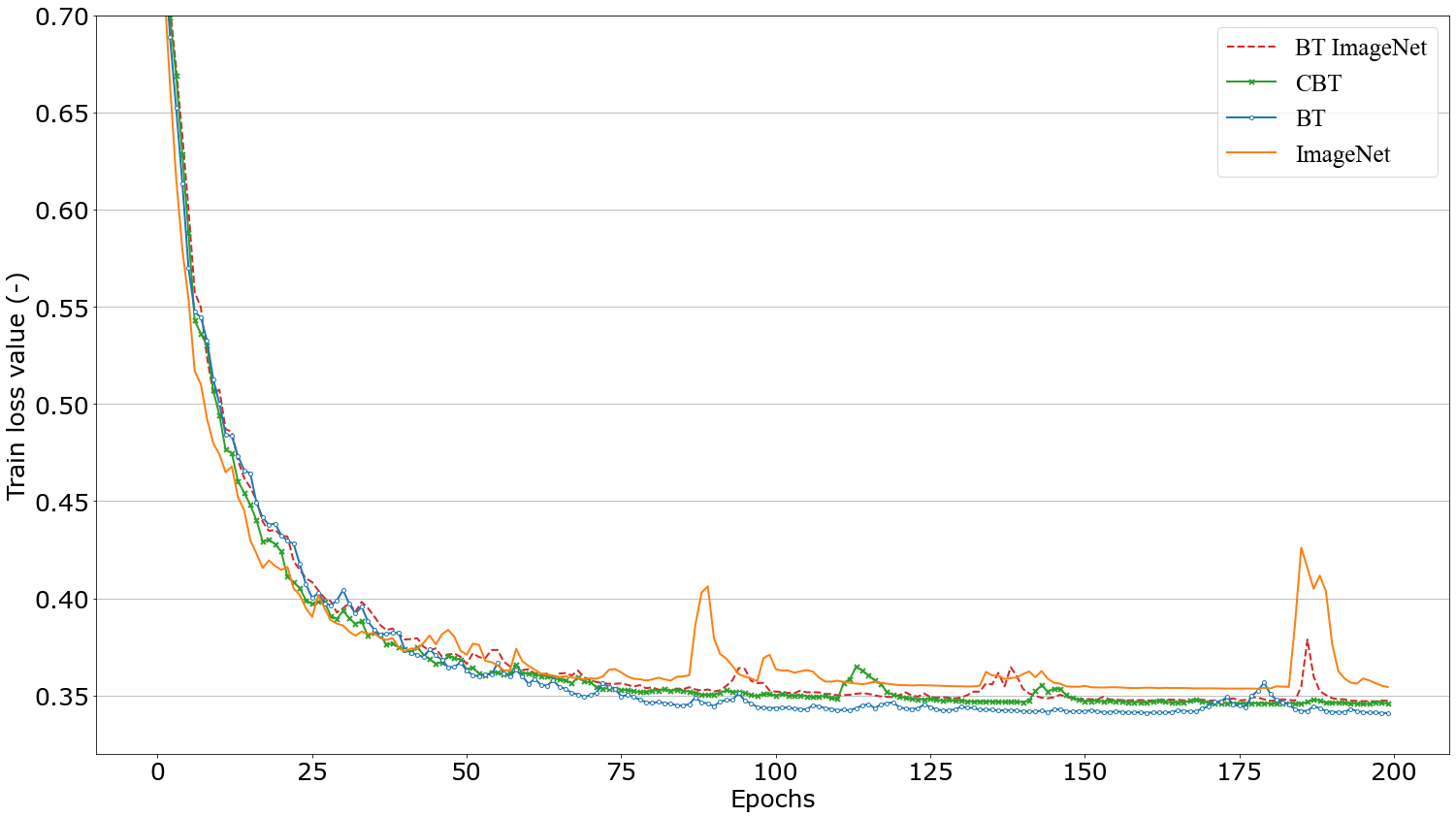}
  \caption{Value of the loss on experiments with 10\% labeled data of respectevely a) UAVid, b) US3D and c) Potsdam.}
  \label{fig:loss_10}
\end{figure}
%
The results of the experiments on the downstream task are shown both in Figure \ref{fig:miou} and Tables \ref{tab:uav}, \ref{tab:pot}, \ref{tab:us3d}. As can be seen, the feature extractors that perform best are the ones obtained from the CBT and BT training on the three selected EO datasets. Precisely, these conformations outperform, with reference to mIoU, their counterpart trained with ImageNet supervised pretraining respectively by 3.39\% and by 3.69\% in average. 
Moreover, it is interesting to note that the performance of the models with the CBT feature extractor is only slightly lower (average drop of a negligible $\approx$0.3\%, referring to mIoU) than that obtained with an encoder trained by means of BT, demonstrating how the proposed approach gives up only a slight part of optimal performance, against clear advantages in terms of computational efficiency and general versatility. In absolute terms, it is necessary to notice once again how self-supervision strategies lead to better results than exclusively supervised ones \cite{stojnic2021self, ayush2021geography}, and, above all, how this is even more true when combining EO data from domains that are not homogeneous in terms of type of sensor, acquisition, resolution and objects represented.
In particular, in the next paragraphs, we will go in the depth of some specific evidences regarding: computational times (Sec. \ref{sec:times}), UAVid experiments (Sec. \ref{sec:res_uav}), Potsdam experiments (Sec. \ref{sec:res_pot}),  US3D experiments (Sec. \ref{sec:res_us}) and catastrophic forgetting (Sec. \ref{sec:cat_forg}).

\subsection{Computational times}
\label{sec:times}

As stated earlier, one of the best advantages of the proposed new method is the shorter computational time with a very limited performance drop in the case of incremental data availability. As already stressed, this situation is especially likely in the field of EO, where new data often arrives continuously \cite{tasar2019incremental}, due to satellite revisit times, scheduled acquisition campaigns and other variable parameters. In Table \ref{tab:times} we can observe the results of the experiments. Concerning the traditional BT strategy, for the training of the three considered datasets, we simulate an incremental arrival of data as follows:
\begin{enumerate}[(1)]

    \item training of the only US3D;
    \item joint training of US3D and UAVid;
    \item training of the three datasets together.
\end{enumerate}
This strategy is employed to create a salient feature extractor for all datasets, with the mean of avoiding catastrophic forgetting, in situations of incremental data availability. 
On the other hand, for CBT, step 1) is referred to training on US3D, step 2) on UAVid and step 3) on Potsdam, as previously stated. We can easily affirm, observing Table \ref{tab:times}, that our method could save nearly 50\% of times, when the data are available incrementally. It is also interesting to state that, also in case of complete and immediate availability of all data, the computational times are comparable (28.75 h for CBT vs 24.61 h for BT, where the computational times of the latter consist of just the third step).

\begin{table}
\centering
\caption{Elapsed training times. CBT offers important advantages when data are provided in an incremental fashion.}
\begin{tabular}{ccccc}
\hline
Strategy & Step 1 (s) & Step 2 (s) & Step 3 (s) & Tot (h) \\ \hline
BT & 41400 & 61500 & 88600 & 53.19 \\
CBT & 42000 & 36400 & 25100 & 28.75 \\ \hline
\end{tabular}
\label{tab:times}
\end{table}

\subsection{UAVid}
\label{sec:res_uav}
According to the results shown in Table \ref{tab:uav} and represented in Figures \ref{fig:miou} a) and \ref{fig:loss_10} a), when using a limited amount of data, the performance of the supervised pretrained encoder are inferior overall. Looking at the Figure \ref{fig:loss_10} a), we can see that a better encoder, when 10\% are used, leads to a more stable and effective training. On the other hand, the training with 50\% of data are the most similar along the different pretrained encoders, with just $\sim$0.5\% mIoU gap between the the worst result (74.92\% mIoU, obtained with ImageNet encoder) and the best (75.39\% mIoU, achieved with BT encoder), that it is almost negligible considering the standard deviations of the results. This trend can be mainly explained by what has been stated above with respect to the image domain. In fact, the images captured by drone are definitely more similar to close-range camera taken images, like ImageNet ones, than those from the other two RS datasets. This is mainly due to the point of view from which the images were captured. For Potsdam and US3D, the viewpoint is almost nadiral, while for UAVid it is oblique, more precisely the camera angle is set to around 45 degrees to the vertical direction, at a flight height of about 50m. As also stated by the authors \cite{lyu2020uavid}, a non-nadiral view allows easier reconstruction of object geometry (i.e. volume, shape, etc...), making the use of more sophisticated feature extractors less effective. These reasons favour the high performance of ImageNet pretrained models, especially with a limited number of labels. However, by increasing the number of labels, the models with encoders trained on the proposed datasets are able to match their features to the best conformation to solve the task with the best performance (80.64\% mIoU with respect to 77.87\% mIoU of the ImageNet encoder experiment). In addition, the fact that the pretraining on UAVid was the second of the three steps slightly affected the performance of the CBT pretraining strategy (80.12\% mIoU), with a very limited drop in performance ($\sim$0.5\%).


\begin{table}[]
\centering
\caption{UAVid results for different \% of training data. The highest score is marked in bold. The second highest is underlined. The tab reports the mean and the standard deviation of three experiments.}
\label{tab:uav}
\begin{tabular}{cccc}
\hline
\textbf{Encoder} & \textbf{\%} & \textbf{mIoU} & \textbf{F1} \\ \hline
 & 10\% & 69.79 ± 0.48 & 80.32 ± 0.34 \\
ImageNet & 50\% & 74.92 ± 0.42 & 84.60 ± 0.28 \\
 & 100\% & 77.87 ± 0.37 & 86.67 ± 0.25 \\ \hline
 & 10\% & 69.99 ± 0.47 & 81.33 ± 0.32 \\
BT ImageNet & 50\% & 75.17 ± 0.39 & 84.81 ± 0.28 \\
 & 100\% & 78.51 ± 0.38 & 86.70 ± 0.26 \\ \hline
 & 10\% & 71.48 ± 0.47 & 81.50 ± 0.31 \\
CBT & 50\% & 75.25 ± 0.42 & 84.73 ± 0.29 \\
 & 100\% & \underline{80.12 ± 0.39} & \underline{88.42 ± 0.26} \\ \hline
 & 10\% & 71.60 ± 0.48 & 81.64 ± 0.35 \\
BT & 50\% & 75.39 ± 0.41 & 84.85 ± 0.27 \\
 & 100\% & \textbf{80.64 ± 0.37} & \textbf{88.44 ± 0.25} \\ \hline
\end{tabular}
\end{table}


\subsection{Potsdam}
\label{sec:res_pot}
As far as the results on Potsdam are concerned, in Table \ref{tab:pot} and Figures \ref{fig:miou} c) and \ref{fig:loss_10} c), we see that the gap between results with self-supervised (71.42\% mIoU) and supervised encoder (64.12\% mIoU) is the largest among all experiments. This trend is true also for the experiments with a limited amount of training data, as the gap in the curve of Figure \ref{fig:loss_10} c) shows. For example, with 10\% of data, the gap between the ImageNet encoder (58.36\% mIoU) and BT encoder (62.72\% mIoU) is $\sim$4.4\%. This can be explained by the fact that this dataset is the one with the least amount of data among the three available (see Table \ref{tab:datasets}). This insight is supported by the fact that the gap between performance with ImageNet encoders and performance with CBT and BT encoders is wider also for the other datasets when only 10\% of the data is used (see also Figure \ref{fig:miou}). Therefore, it is definitely the one that benefits the most from a more efficient encoder feature selection. This could be confirmed by the fact that the Potsdam domain is comparable with that of US3D, a very wide dataset, capable of improving the representations that can be used during the training of the Potsdam semantic segmentation, confirming similar intuitions reached, for example, in \cite{Reed_2022_WACV}.


\begin{table}[]
\centering
\caption{Potsdam results for different \% of training data. The highest score is marked in bold. The second highest is underlined. The tab reports the mean and the standard deviation of three experiments.}
\label{tab:pot}
\begin{tabular}{cccc}
\hline
\textbf{Encoder} & \textbf{\%} & \textbf{mIoU} & \textbf{F1} \\ \hline
 & 10\% & 58.36 ± 0.63 & 70.73 ± 0.50 \\
ImageNet & 50\% & 61.59 ± 0.57 & 73.55 ± 0.44 \\
 & 100\% & 64.12 ± 0.54 & 75.98 ± 0.40 \\ \hline
 & 10\% & 61.45 ± 0.59 & 73.25 ± 0.51 \\
BT ImageNet & 50\% & 66.88 ± 0.55 & 77.41 ± 0.42 \\
 & 100\% & 70.22 ± 0.52 & 79.42 ± 0.39 \\ \hline
 & 10\% & 60.90 ± 0.62 & 72.46 ± 0.52 \\
CBT & 50\% & 67.24 ± 0.55 & 77.87 ± 0.44 \\
 & 100\% & \underline{70.90 ± 0.52} & \underline{80.01 ± 0.39} \\ \hline
 & 10\% & 62.72 ± 0.60 & 74.44 ± 0.51 \\
BT & 50\% & 67.29 ± 0.54 & 77.77 ± 0.43 \\
 & 100\% & \textbf{71.42 ± 0.52} & \textbf{80.63 ± 0.39} \\ \hline
\end{tabular}
\end{table}


\subsection{US3D}
\label{sec:res_us}
As far as the US3D dataset is concerned, again self-supervision techniques lead to better results on downstream tasks, highlighting how, in this case, CBT performs best of all (CBT 84.56\% vs ImageNet 83.41\% mIoU). In fact, this is the first evidence that can be easily argued from Table \ref{tab:us3d} and Figures \ref{fig:miou} b) and \ref{fig:loss_10} b). Therefore, considered also the standard deviations of the final results, we observe that there are no significant differences in performance between the other encoders (BT ImageNet 84.49\% vs CBT 84.56\% vs BT 84.43\% mIoU), since the US3D is a large dataset, composed of several images of the same area but captured from different points of view (i.e. multiview). This redundancy, working as data augmentation itself, facilitates the resolution of the task on this dataset, as once an efficient feature extractor is engaged, convergence is achieved quite effectively. This intuition is confirmed also by the training curves, showed in Figure \ref{fig:loss_10} b), where the training, except of some small instability in ImageNet curve, follow a similar behavior. It is not surprising that similar results are presented in \cite{stojnic2021self}, where self-supervision is applied on other large datasets.


\begin{table}[]
\centering
\caption{US3D results for different \% of training data. The highest score is marked in bold. The second highest is underlined. The tab reports the mean and the standard deviation of three experiments.}
\label{tab:us3d}
\begin{tabular}{cccc}
\hline
\textbf{Encoder} & \textbf{\%} & \textbf{mIoU} & \textbf{F1} \\ \hline
 & 10\% & 74.18 ± 0.41 & 84.60 ± 0.26 \\
ImageNet & 50\% & 78.25 ± 0.37 & 87.57 ± 0.14 \\
 & 100\% & 83.41 ± 0.30 & 90.76 ± 0.12 \\ \hline
 & 10\% & 75.44 ± 0.39 & 85.38± 0.27 \\
BT ImageNet & 50\% & 79.26 ± 0.35 & 87.89 ± 0.13 \\
 & 100\% & 84.49 ± 0.28 & 91.27 ± 0.11 \\ \hline
 & 10\% & 75.31 ± 0.41 & 85.42 ± 0.27 \\
CBT & 50\% & 79.70 ± 0.36 & 88.20 ± 0.14 \\
 & 100\% & \textbf{84.56 ± 0.28} & \underline{91.28 ± 0.12} \\ \hline
 & 10\% & 75.89 ± 0.39 & 85.69 ± 0.25 \\
BT & 50\% & 80.67 ± 0.36 & 88.94 ± 0.13 \\
 & 100\% & \underline{84.43 ± 0.28} & \textbf{91.30 ± 0.12} \\ \hline
\end{tabular}
\end{table}


\subsection{Overcoming Catastrophic Forgetting}
\label{sec:cat_forg}

CL strategy, in addition to generically improving performance as demonstrated in previous sections, has the main advantage of overcoming catastrophic forgetting. To illustrate this, we performed a series of experiments in which we trained a BT model in a sequential fashion, without introducing any CL strategy. Specifically, starting from a ResNet50 pretrained on ImageNet with BT, we performed three training steps, in which we used the model obtained in the previous step: i) BT on US3D dataset; ii) BT on UAVid; iii) BT on Potsdam. Finally, we used the resulting ResNet50 as the backbone of UNet++ model, for the semantic segmentation downstream task. Figure \ref{fig:cat_for} and Table \ref{tab:cat_for} show the effectiveness of CBT as a strategy to overcome catastrophic forgetting, making possible to train a powerful encoder, without the need of relying on either all the data together or high computational resources. Particularly, we can affirm that constraining the parameters of the model pretrained on ImageNet is an easy and effective strategy to train the backbone. In fact, when it is not possible to rely on a vast amount of data specifically shaped for EO tasks, it is better to exploit the capabilities of pretrained models that used huge dataset, like in this scenario.

\begin{figure}
  \includegraphics[width=1.\linewidth]{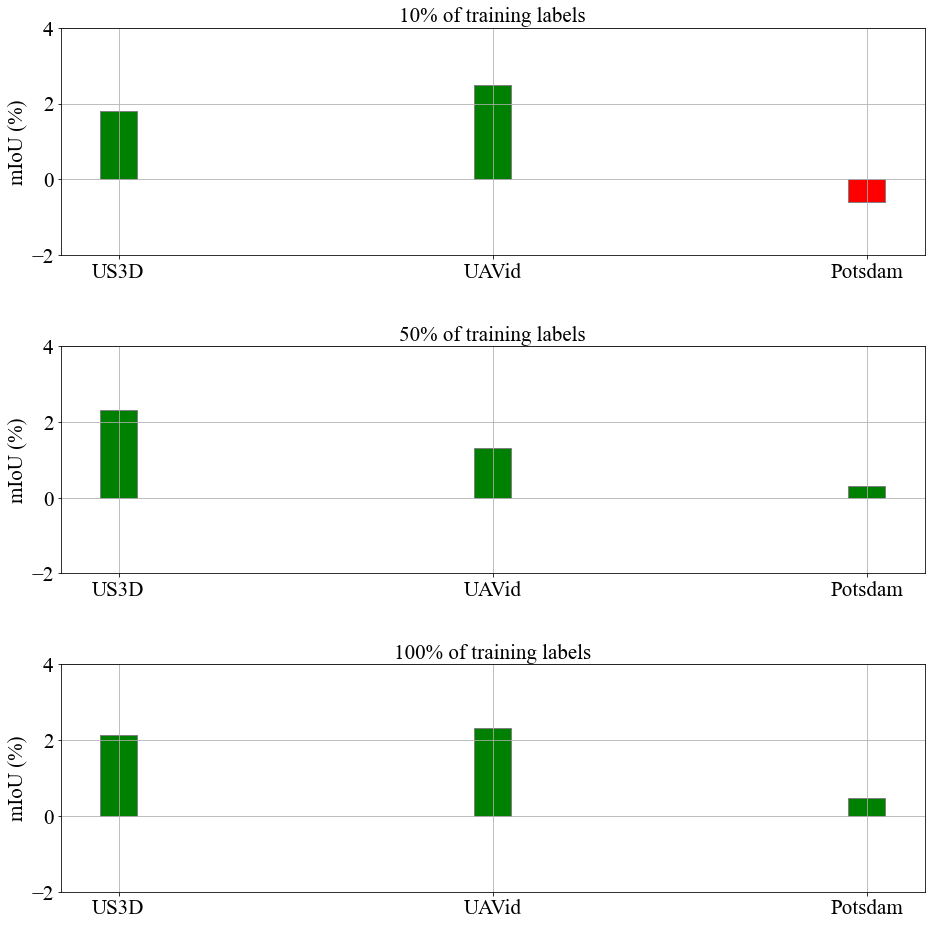}
  \caption{differences of mIoU among the experiments obtained with the encoder pretrained with the proposed CBT and the catastrophic forgetting baseline (i.e. encoder pretrained with a vanilla BT sequentially trained on the three datasets).}
  \label{fig:cat_for_diff}
\end{figure}

\begin{table}[]
\centering
\caption{mIoU values for experiments with different amounts of labeled data with the encoder pretrained with: i) CBT, ii) a vanilla BT trained consecutively on the three datasets.}
\label{tab:cat_for}
\begin{tabular}{ccccc}
\hline
\multirow{2}{*}{Dataset} & \multirow{2}{*}{Encoder} & \multicolumn{3}{c}{mIoU (\%)} \\ \cline{3-5} 
 &  & 10\% & 50\% & 100\% \\ \hline
\multirow{2}{*}{US3D} & CBT & 75.31 & 79.70 & 84.56 \\
 & BT & 73.50 & 77.39 & 82.41 \\ \hline
\multirow{2}{*}{UAVid} & CBT & 71.48 & 75.25 & 80.12 \\
 & BT & 68.98 & 73.93 & 77.81 \\ \hline
\multirow{2}{*}{Potsdam} & CBT & 60.90 & 67.24 & 70.90 \\
 & BT & 61.51 & 66.94 & 70.42 \\ \hline
\end{tabular}
\end{table}

Moreover, we can see in Figure \ref{fig:cat_for} and Table \ref{tab:cat_for}, that once again UAVid is the dataset that, being more different from the others, suffer most from catastrophic forgetting (e.g. drop of $\sim2.5\%$ when $10\%$ of labels are used, $\sim2\%$ when $100\%$ of labels). In fact, as we have already observed, Potsdam and US3D have both nadiral views, making their characteristics more similar. For this very reason, the performance of 3-step BT on US3D is never excessively worse than the counterpart trained with CBT, even though the average performance drop (of $\sim1.5\%$) is not negligible, being the first of the three dataset used. On the other hand, as one can expect, the performance on the Potsdam dataset, with the 3-step BT, are really similar to the one reached with CBT. In fact, being the last dataset on which the algorithm is trained, most of the knowledge of the encoder came from this dataset. This is visible especially when few data are used, where 3-step BT performance (61.51\% mIoU) overcome the CBT one (60.90\% mIoU). In general, once again, given the required computational power and the overall performance, CBT seems the best solution to have consistent results on all the datasets.



\section{Conclusions}
\label{sec:conc}

In this paper, we have shown that the combination of CL and SSL offers an optimal compromise between performance and training efficiency and versatility for RS applications. In particular, we demonstrated a combined approach (Continual Barlow Twins) leading to consistent performance in a novel combination of datasets with RS images that are heterogeneous in terms of sensors, resolution, acquisition and scenes represented. Since the availability of unlabeled data is increasing at a great speed, and it is not possible for everyone to train repeatedly large models, a framework like CBT offers a potential solution. However, more work remains to be done. First, the validity of these results could be extended to new datasets and new tasks. Among the use of new datasets, we mention datasets containing multispectral images (i.e., not only with RGB bands). Second, other SSL and CL strategies can be combined into an effective and efficient framework.

\bibliographystyle{ieeetr}
\bibliography{refs}

\end{document}